\documentclass[sigconf,nonacm]{acmart}

\renewcommand\footnotetextcopyrightpermission[1]{}
\setcopyright{none}

\usepackage{booktabs}
\usepackage{array}
\usepackage{tabularx}
\usepackage{enumitem}
\usepackage{listings}
\usepackage{xcolor}
\usepackage{balance}
\usepackage{url}

\definecolor{codegray}{RGB}{245,245,245}
\providecommand{\tightlist}{%
  \setlength{\itemsep}{0pt}\setlength{\parskip}{0pt}}

\title{Executable Code Knowledge: Code as a Native, Validation-Carrying Knowledge Representation for AI Coding Agents}

\author{Xueping Gao}
\affiliation{%
  \institution{Alibaba Cloud}
  \city{Hangzhou}
  \country{China}}
\email{xueping.gxp@alibaba-inc.com}

\begin{document}

\begin{abstract}
AI coding agents need more than relevant snippets: they need business semantics,
validation evidence, relations, and assurance that their context is current.
Existing systems usually infer or externalize this knowledge through retrieval,
summaries, graphs, rules, or reverse specifications. We investigate a
complementary representation in which selected code units directly carry
agent-usable knowledge. We introduce Executable Code Knowledge (ECK) and
define an Executable Code Knowledge Unit (ECKU) as a source-bound object
combining stable identity, semantics, executable behavior, contracts, evidence,
relations, provenance, validation state, and a query interface. Our Python
prototype supports code-local authoring, manifest export, evidence execution,
exact changed-line impact, freshness checking, and agent-facing projections.
Across three real Python repositories and 26 controlled patch tasks, direct ECK
provides executable test coverage for 11/11 evidence-bearing tasks and exact
selectors for 9/11; hiding declared evidence reduces exact recovery to 1/11
(paired exact McNemar p=0.0078). ECK-derived rules recover 11/11 exact selectors,
showing that rules are effective delivery artifacts while ECK supplies source
binding, validation state, impact, and freshness. Exact changed-line impact
matches independently authored labels on all 26 patches (12 unit links;
precision, recall, and F1 all 1.000). AST-bounded fingerprints classify 50
positive changes and 17 unrelated same-file controls correctly, whereas static
rules snapshots detect none of the 50 stale cases. Model-backed patch-review
and cross-layer studies measure projection fidelity rather than independent
impact discovery. These results support a hybrid architecture: retrieval for
coverage, ECK for source and evidence governance, and projections for delivery.
\end{abstract}


\maketitle

\section{Introduction}\label{introduction}

AI coding agents are increasingly evaluated on repository-level issue resolution, code retrieval, tool use, and test-based validation. This shift has exposed a gap in how software knowledge is represented. Agents do not only need snippets that look similar to a request. They need to know which behavior is a business rule, which tests are authoritative, which API or data relation is affected, and whether the context they are using is still fresh.

Most current systems construct such knowledge from code after the fact:

\begin{lstlisting}
ordinary code -> retrieval / summary / static graph / reverse spec -> agent context
\end{lstlisting}

This extractive pipeline is useful and often necessary for coverage. However, it has an authority problem. Retrieved chunks can suggest where to look, but they usually do not state which evidence validates a rule. Static graphs expose structure, but they do not say whether a relation is a domain invariant or an incidental call edge. Generated summaries and reverse-engineered specifications can be helpful, but their claims must still be traced, validated, and kept fresh.

This paper asks a different question:

\begin{quote}
Can code itself directly represent knowledge for AI coding agents, rather than serving only as text to be mined into external knowledge objects?
\end{quote}

We answer yes for a specific class of high-value code units. We propose \textbf{Executable Code Knowledge (ECK)}: a representation in which selected implementation units directly carry domain semantics, executable behavior, contracts, evidence, relations, provenance, validation state, and query operations. The unit of representation is an \textbf{Executable Code Knowledge Unit (ECKU)}.

ECK is not intended to annotate every function or replace retrieval over the whole repository. Instead, it targets code units where correctness and context are especially valuable: business rules, policies, security checks, data transformations, API behavior, parser invariants, workflow transitions, and compatibility contracts. The practical thesis is:

\begin{quote}
Broad extractive methods provide coverage; direct executable code knowledge provides source-bound identity, declared executable support, and freshness for selected high-value units.
\end{quote}

\subsection{Contributions}\label{contributions}

This paper makes four contributions:

\begin{enumerate}
\def\labelenumi{\arabic{enumi}.}
\item
  \textbf{Problem framing.} We formulate direct code knowledge representation for AI coding agents: code units should be able to express not only computation, but also machine-usable semantics, evidence, relations, and freshness.
\item
  \textbf{Representation.} We define ECKUs as code-authored executable knowledge objects:

  \begin{lstlisting}
  ECKU = <I, S, B, C, E, R, P, V, Q>
  \end{lstlisting}

  where the fields represent identity, semantics, executable body, contract, evidence, relations, provenance, validation state, and query interface.
\item
  \textbf{Prototype and artifact.} We implement ECK for Python with decorators, manifest generation, evidence validation, direct source-span impact, AST-bounded freshness, candidate migration, agent context generation, and machine- or reviewer-readable patch evidence reports.
\item
  \textbf{Mechanism-level evaluation.} We evaluate declared-evidence recovery, exact changed-line impact against independent manual labels, deterministic impact-report consumption, projection fidelity, and freshness sensitivity/specificity. The strongest result is not generic localization: direct ECK turns selected code units into validation-carrying and freshness-checkable knowledge objects, while rules remain an effective delivery format.
\end{enumerate}

\section{Motivation}\label{motivation}

Consider a pricing rule:

\begin{lstlisting}
def vip_large_order_discount(user, order):
    return Discount(rate=0.9)
\end{lstlisting}

An agent can inspect this function and infer that VIP users may get a 10\% discount. But it must still guess:

\begin{itemize}
\tightlist
\item
  whether this is a business rule or an incidental implementation detail;
\item
  which preconditions make the rule valid;
\item
  which endpoint or data fields the rule belongs to;
\item
  which tests should be run before modifying it;
\item
  whether previously generated documentation or summaries are still fresh.
\end{itemize}

An ECKU expresses this information directly:

\begin{lstlisting}
@knowledge_unit(
    id="pricing.discount.vip_large_order",
    domain="pricing",
    intent="VIP users receive a 10% discount for orders above 100.",
    semantics={
        "business_rule": "VIP large-order discount",
        "threshold": 100,
        "discount_rate": 0.9,
    },
    relates_to={
        "api": ["POST /orders/quote"],
        "data": ["users.level", "orders.amount"],
        "policy": ["pricing.discount_policy"],
    },
)
@evidence(
    id="tests/test_pricing.py::test_vip_large_order_discount",
    kind="pytest",
    command="python -m pytest tests/test_pricing.py::test_vip_large_order_discount",
    claim="The executable test checks the VIP large-order discount rate.",
)
@contract(
    requires=["user.level == 'VIP'", "order.amount > 100"],
    ensures=["result.rate == 0.9"],
)
def vip_large_order_discount(user: User, order: Order) -> Discount:
    return Discount(rate=0.9)
\end{lstlisting}

The key distinction is source binding with executable support. This is not a summary generated from code. It is a code-authored assertion whose declared evidence can be executed and whose freshness can be checked; it is not assumed to be semantically true merely because it was annotated.

\section{Related Work}\label{related-work}

\subsection{Repository-Level Code Retrieval and Context Benchmarks}\label{repository-level-code-retrieval-and-context-benchmarks}

Recent benchmarks argue that agentic coding requires repository-level search rather than isolated snippet retrieval. ContextBench \citep{contextbench2026} introduces a process-oriented evaluation of context retrieval in coding agents with human-annotated gold contexts across 1,136 issue-resolution tasks from 66 repositories, measuring recall, precision, and efficiency during agent trajectories. CORE-Bench \citep{corebench2026} similarly frames agentic coding as requirement-driven repository search and evaluates code understanding, issue-to-edit localization, and broader context retrieval over large query and relevance-label sets.

ECK is not a competing retrieval benchmark. These works clarify why context representation matters. Our contribution is complementary: rather than only retrieving more relevant context, we ask whether selected code units can carry validation-ready knowledge before retrieval occurs.

\subsection{Code Graphs and Graph-Based Agent Interfaces}\label{code-graphs-and-graph-based-agent-interfaces}

CodexGraph \citep{codexgraph2025} integrates LLM agents with graph database interfaces extracted from code repositories, enabling graph queries over repository structure. Sourcegraph and related code-intelligence products similarly emphasize repository-wide indexing, code search, cross-repository context, navigation, and MCP-style access for agents. Code KG and Graph-RAG systems transform repositories into structured external representations for code navigation and generation.

ECK differs in where knowledge identity and validation provenance reside. In code graph and code-search systems, the graph or index is extracted from ordinary code and maintained as an external interface. In ECK, the source-bound knowledge object is authored with the code unit itself; manifests, graph-like relations, rules, and context packs are projections from that code-authored unit.

\subsection{Rules, Memories, and Repository Instruction Files}\label{rules-memories-and-repository-instruction-files}

Modern coding tools increasingly support persistent project context. Cursor supports project, team, and user rules as well as \path{AGENTS.md}; Devin and Windsurf-style workflows emphasize memories or knowledge bases; GitHub Copilot and VS Code support custom instruction files; and \path{AGENTS.md} \citep{agentsmd} has emerged as a simple repository-local format for setup commands, coding conventions, and agent guidance. These systems address a practical problem: agents need stable project-specific instructions instead of relying only on ad hoc prompts.

ECK is compatible with this direction but makes a different technical claim. Rules and memories are effective \textbf{delivery artifacts}: they can tell an agent what to do, and our Rules Context baseline confirms that fresh rules can match ECK on validation-command recovery. Their weakness is provenance and maintenance. A Markdown rule can mention an identifier and validation command, but it is not normally backed by a source span, contract hash, evidence fingerprint, validation state, or patch-impact operation. ECK treats rules and memories as projections from a source-bound object, so the system can ask whether the projected guidance is still fresh and which declared unit a patch directly overlaps.

\subsection{Reverse Documentation and Operational Specifications}\label{reverse-documentation-and-operational-specifications}

Reversa \citep{reversa2026} converts legacy software into traceable operational specifications for AI agents using a multi-agent reverse documentation pipeline. It emphasizes traceability, confidence marking, and preservation of gaps for human validation.

ECK shares the goal of agent-usable operational knowledge, but differs in direction. Reversa extracts or reconstructs specifications from existing systems. ECK proposes an authoring model in which selected implementation units directly carry their operational knowledge, evidence, and freshness state. In practice the two can be combined: reverse engineering can propose candidate ECKUs, while human-reviewed ECKUs become authoritative code-native knowledge.

\subsection{Executable Memory and Code as Agent Infrastructure}\label{executable-memory-and-code-as-agent-infrastructure}

\textbf{User as Code (UaC)} \citep{uac2026} is the closest recent statement of the representation thesis: it checkpoints conversational facts into typed Python state and executable rules so aggregation and policy execution become ordinary computation rather than retrieval. ECK applies a narrower idea to software engineering artifacts. Its state is not a generated user model; an ECKU is a reviewed implementation unit bound to a repository source span, contract, validation evidence, patch impact, and post-validation freshness.

\textbf{Metis} \citep{metis2026} provides a controlled comparison of text and code memory and finds complementary trade-offs, then crystallizes repeated experience into validated callable tools. This supports rather than contradicts our negative result that raw ECK need not be the best delivery format. Metis represents agent experience for reuse across interactive tasks; ECK represents selected repository behavior and projects it into rules or typed patch reports while retaining source and validation provenance.

\textbf{Code as Agent Harness} \citep{codeharness2026} generalizes code from an output medium to infrastructure for reasoning, action, memory, coordination, and execution-based verification. ECK is a concrete, deliberately smaller realization within repository-level coding: a code-native knowledge and evidence interface for patch preparation and review, rather than a general agent harness architecture.

\subsection{Governance, Skills, and Persistent Code Graphs}\label{governance-skills-and-persistent-code-graphs}

\textbf{Protocol-Driven Development (PDD)} \citep{pdd2026} is the strongest conceptual foil. It makes a machine-enforceable protocol of structural, behavioral, and operational invariants sovereign, treats code as a replaceable realization, and records an evidence chain for admission. ECK makes a different authority choice: selected implementation units are the source-bound assertions from which agent-facing projections are generated. ECK does not define the full admissible implementation space, and passing declared evidence is not a proof of semantic truth. The common ground is continuous invariants, provenance, and executable evidence for generated software.

\textbf{SWE-Skills-Bench} \citep{sweskills2026} shows that procedural skill packages often provide little benefit and can degrade performance when guidance is version-mismatched. Its end-to-end acceptance tests are stronger than our planning and projection evaluations. The version-mismatch result directly motivates ECK freshness, while also setting a future bar: show that fresh, source-bound evidence improves patch acceptance rather than only evidence recovery.

\textbf{Codebase-Memory} \citep{codebasememory2026} constructs a persistent Tree-Sitter knowledge graph with call traversal, impact analysis, community discovery, and an MCP interface across many languages and repositories. It is substantially broader than ECK for structural exploration and transitive impact. ECK is narrower and currently Python-specific: its contribution is author-declared domain semantics, contracts, executable evidence, and validation freshness for selected units. We therefore claim exact direct source-span impact, not a replacement for graph-based relation expansion.

\subsection{Design by Contract, Literate Programming, and Documentation as Code}\label{design-by-contract-literate-programming-and-documentation-as-code}

Design by Contract expresses preconditions, postconditions, and invariants \citep{meyer1992}. Literate programming keeps explanation and programs close in a reviewable source artifact \citep{knuth1984}. Dynamic invariant detection such as Daikon instead infers likely specifications from executions \citep{ernst2001}, illustrating the complementary extraction path in our coverage-authority frontier. Documentation-as-code practices preserve prose and executable examples in version control and CI.

ECK incorporates contracts and executable evidence, but its target is different: structured, queryable, freshness-aware knowledge objects for agents. Contracts are one field in an ECKU, not the full representation. Prose documentation can explain behavior, but it usually lacks a typed agent interface, patch impact mapping, and persisted validation fingerprints.

ECK's direct span intersection also belongs to the broader change-impact and regression-test-selection lineage \citep{bohner1996, rothermel1997}. Classical impact analysis asks which artifacts may be affected by a change, while safe regression-test selection reasons about which tests must be rerun. Requirements traceability similarly links requirements to downstream artifacts \citep{gotel1994}. Our prototype makes a narrower claim: for selectively authored units, an exact changed line can be mapped to a declared knowledge identity and its evidence. It does not replace dependency-based impact propagation, safe test selection, or requirements traceability.

\section{Executable Code Knowledge}\label{executable-code-knowledge}

We define an Executable Code Knowledge Unit as:

\begin{lstlisting}
ECKU = <I, S, B, C, E, R, P, V, Q>
\end{lstlisting}

where:

\begin{itemize}
\tightlist
\item
  \path{I}: stable identity;
\item
  \path{S}: domain semantics;
\item
  \path{B}: executable body;
\item
  \path{C}: contract;
\item
  \path{E}: evidence;
\item
  \path{R}: relations to APIs, data entities, tests, documents, or other units;
\item
  \path{P}: provenance;
\item
  \path{V}: validation state and freshness;
\item
  \path{Q}: query and composition interface.
\end{itemize}

\subsection{Validation and Freshness}\label{validation-and-freshness}

For an ECKU \path{u}, define:

\begin{lstlisting}
SourceHash(u)    = H(source(u.B))
KnowledgeHash(u) = H(u.I, u.S, u.R, relevant(u.P))
ContractHash(u)  = H(u.C)
EvidenceHash(u)  = H(u.E)

Fresh(u, t) iff
  SourceHash_t(u)    = SourceHash_last_validated(u)
  KnowledgeHash_t(u) = KnowledgeHash_last_validated(u)
  ContractHash_t(u)  = ContractHash_last_validated(u)
  EvidenceHash_t(u)  = EvidenceHash_last_validated(u)

Executable(u) = {e in u.E | e declares an executable command}

Supported(u) iff Executable(u) is non-empty and
                 for every e in Executable(u), Run(e) passes

Valid(u) iff Fresh(u) and Supported(u)
\end{lstlisting}

This conservative all-evidence aggregation matches the prototype. It does not claim that passing tests prove semantic truth; it records that all executable support declared by the unit currently passes. Claim-level required/supporting evidence is a natural extension. The model deliberately separates relevance from validity: a retrieved code chunk may be relevant, while an ECKU additionally carries declared executable support and a freshness state.

\subsection{Query and Composition}\label{query-and-composition}

An ECK runtime should expose:

\begin{lstlisting}
build(repo)
query(repo, intent | symbol | relation | evidence | domain)
show(unit_id)
validate(unit_id)
freshness(unit_id)
impact(patch)
context(task)
\end{lstlisting}

Composition can be relation-based: API endpoint to business rule, business rule to data fields, code unit to tests, policy to multiple rules. Our prototype implements build, query, validation, freshness checking, direct patch impact, context generation, and patch evidence reporting; relation-based \path{compose} remains future work.

\section{System Design and Implementation}\label{system-design-and-implementation}

The Python prototype contains five layers.

\subsection{Authoring Layer}\label{authoring-layer}

Developers author ECKUs with decorators:

\begin{itemize}
\tightlist
\item
  \path{@knowledge_unit(...)}
\item
  \path{@contract(...)}
\item
  \path{@evidence(...)}
\end{itemize}

The implementation preserves the original executable function as the source span even when contract wrappers are applied. This matters because an agent should see provenance for the business function, not for a generated wrapper.

\subsection{Discovery and Manifest Layer}\label{discovery-and-manifest-layer}

The builder imports repository modules, discovers registered ECKUs, records source spans and source hashes, and exports:

\begin{lstlisting}
.eck/knowledge_units.jsonl
.eck/evidence.jsonl
.eck/validation_state.json
\end{lstlisting}

\subsection{Validation Layer}\label{validation-layer}

Validation executes evidence commands and records:

\begin{itemize}
\tightlist
\item
  evidence status;
\item
  source hash;
\item
  contract hash;
\item
  evidence hash;
\item
  stale reasons;
\item
  validation timestamp.
\end{itemize}

\subsection{Freshness Layer}\label{freshness-layer}

Freshness checking compares executable-body, agent-facing knowledge, contract, and evidence fingerprints with the persisted validation state. \path{KnowledgeHash} covers stable identity, symbol, domain, intent, semantics, normalized relations, and relevant authoring provenance; relation order is canonicalized. Function boundaries come from the Python AST (\path{lineno}/\path{end_lineno}), and the source fingerprint excludes decorators so knowledge, contract, and evidence changes retain field-specific stale reasons. The checker reports \path{never_validated}, \path{source_changed_since_validation}, \path{knowledge_changed_since_validation}, \path{contract_changed_since_validation}, and \path{evidence_changed_since_validation}.

\subsection{Agent Context Layer}\label{agent-context-layer}

The context command emits compact typed context:

\begin{lstlisting}
python -m eck context <repo> "change request"
\end{lstlisting}

The resulting pack includes intent, symbol, source span, contracts, executable evidence commands, and relations. Unlike a generated summary, it does not require the agent to infer which command validates a knowledge claim.

\section{Evaluation}\label{evaluation}

We evaluate four research questions.

\textbf{RQ1: Agent validation planning.} Does direct ECK context help coding agents identify executable validation commands for a planned change?

\textbf{RQ2: Patch impact and validation coverage.} When a patch touches a direct ECKU, can ECK provide impacted knowledge units and validation plans?

\textbf{RQ3: Cross-layer projection.} How faithfully do raw ECK, rules, and task-specific projections transport API, data-field, business-rule, identity, and evidence fields to an agent?

\textbf{RQ4: Freshness.} Can ECK detect when previously validated code knowledge becomes stale after source, contract, or evidence changes?

\subsection{Repositories}\label{repositories}

We use three real Python repositories with local tests and manually authored direct ECKUs:

\begin{table*}[t]
\centering
\caption{Evaluated repositories and controlled patches.}
\label{tab:eck-1}
\scriptsize
\setlength{\tabcolsep}{3pt}
\renewcommand{\arraystretch}{1.05}
\begin{tabularx}{\textwidth}{>{\raggedright\arraybackslash}X>{\raggedleft\arraybackslash}X>{\raggedleft\arraybackslash}X>{\raggedleft\arraybackslash}X}
\toprule
Repository & Direct ECKUs & With Evidence & Controlled Patches \\
\midrule
python-dotenv & 6 & 6 & 10 \\
python-slugify & 5 & 4 & 8 \\
tomli & 6 & 6 & 8 \\
Total & 17 & 16 & 26 \\
\bottomrule
\end{tabularx}
\end{table*}

The ECKUs cover high-value functions such as dotenv variable resolution, slug normalization/truncation, CLI argument behavior, TOML parsing, and parse-float safety.

\subsection{Methods}\label{methods}

We compare:

\begin{itemize}
\tightlist
\item
  \textbf{Plain}: repository tree context only;
\item
  \textbf{BM25 Chunk}: a standard lexical retrieval baseline over fixed-size source chunks;
\item
  \textbf{EKP Retrieval}: extractive knowledge packets derived from code;
\item
  \textbf{Rules Context}: ECK-derived hints rendered as Markdown rules or memories;
\item
  \textbf{Direct ECK without Evidence}: direct ECK context with executable evidence commands hidden;
\item
  \textbf{Direct ECK}: direct code-authored ECKUs retrieved for the task.
\end{itemize}

We evaluate two locally deployed models through a locally hosted vLLM OpenAI-compatible endpoint:

\begin{itemize}
\tightlist
\item
  \path{qwen2.5-7b-instruct};
\item
  \path{qwen2.5-coder-32b-awq}.
\end{itemize}

For each of 26 issue-style change requests, the model outputs a plan with target files, target symbols, validation commands, and rationale. The requests are written independently from patch filenames to reduce answer leakage. Gold labels are derived from the held-out patch touched files, impacted EKPs/ECKUs, and direct ECK evidence commands.

We also run a patch-review workflow. The model receives a patch diff and one of four context policies: patch-only plain context, BM25 source chunks, rules context, or an ECK patch-impact report. It must output affected files, affected ECK IDs, affected symbols, validation commands, risk flags, and rationale. Because the ECK targets and report share the deterministic impact generator, this workflow measures report-consumption and projection fidelity rather than impact-engine accuracy. We report both strict JSON compliance and lenient parseability, because several model outputs use \path{key: value} records rather than valid JSON.

Finally, we run an API-DB-business case study on a synthetic enterprise-style service with three domains: orders/pricing, subscriptions/entitlements, and refunds/risk. The service contains API endpoints, entity/schema fields, cross-layer business rules, and evidence tests across API, service, and schema layers. This case study compares plain, BM25, rules, direct ECK, and an \textbf{ECK projection} condition that renders ECK source-of-truth units into compact agent-facing fields: impacted ECK IDs, APIs, DB fields, business rules, and validation commands.

\subsection{Metrics}\label{metrics}

We report:

\begin{itemize}
\tightlist
\item
  file recall;
\item
  normalized exact-selector recall over tasks with non-empty gold validation hooks;
\item
  executable-coverage recall, where a file- or class-level command covers a more specific gold test;
\item
  patch-review impacted ECK ID recall over ECK-covered patches;
\item
  API, DB-field, and business-rule recall in the cross-layer case study;
\item
  strict JSON and lenient parsed-output rates.
\end{itemize}

The earlier pilot mixed representation IDs with code symbols in its symbol gold; we therefore omit that metric from the main claims. Exact-selector recall requires normalized command and test-selector equality, treating \path{pytest} and \path{python -m pytest} as equivalent launch forms. Executable coverage uses test-file and class/test selector containment; it does not use arbitrary substring matching. ECK-ID and cross-layer recall in the patch-review studies measure projection fidelity because their targets and ECK conditions share the deterministic impact representation.

\section{Results}\label{results}

\subsection{RQ1: Explicit Evidence Controls Precise Validation Selection}\label{rq1-explicit-evidence-controls-precise-validation-selection}

Table 2 shows the main issue-style result with Qwen2.5-Coder-32B-AWQ.

\begin{table*}[t]
\centering
\caption{Validation-planning results with Qwen2.5-Coder-32B-AWQ.}
\label{tab:eck-2}
\scriptsize
\setlength{\tabcolsep}{3pt}
\renewcommand{\arraystretch}{1.05}
\begin{tabularx}{\textwidth}{>{\raggedright\arraybackslash}X>{\raggedleft\arraybackslash}X>{\raggedleft\arraybackslash}X>{\raggedleft\arraybackslash}X}
\toprule
Method & File Recall & Exact Selector Recall* & Executable Coverage Recall* \\
\midrule
Plain & 0.500 & 0.000 & 0.000 \\
BM25 Chunk & 0.981 & 0.455 & 0.455 \\
EKP Retrieval & 0.846 & 0.000 & 0.636 \\
Rules Context & 0.865 & 1.000 & 1.000 \\
Direct ECK without Evidence & 0.981 & 0.091 & 0.636 \\
Direct ECK & 0.981 & 0.818 & 1.000 \\
\bottomrule
\end{tabularx}
\end{table*}

*Metrics are averaged over the 11 tasks with non-empty gold validation hooks. Rules Context exactly recovers all 11 task-level selector sets; Direct ECK exactly recovers 9/11 and supplies commands covering the gold tests for 11/11. BM25 exactly recovers 5/11, while direct ECK without evidence exactly recovers 1/11. This refines the mechanism: executable evidence is especially important for \textbf{precise selector recovery}, while metadata or retrieved source can sometimes suggest a broader file or class that covers the desired test. ECK's additional contribution over rules is not literal delivery accuracy; it is that the evidence is bound to executable code units with source spans, impact analysis, validation state, and freshness checks.

The paired evidence ablation is important. Direct ECK and direct ECK without evidence have identical file recall, but exact task success changes from 9/11 to 1/11, with eight Direct-ECK-only successes and no no-evidence-only success. A two-sided exact McNemar test over the discordant tasks gives p=0.0078125. Coverage recall also drops from 1.000 to 0.636. This supports the narrower mechanism-level claim that explicit evidence primarily improves validation precision.

We also track JSON plan parse failures. In this run, plain context produced 12 unparsable or empty plans, EKP retrieval produced 3, and Rules Context produced 3, while BM25, direct ECK without evidence, and direct ECK produced none. The recall scores count unparsable outputs as empty predictions.

Bootstrap confidence intervals over tasks further clarify the effect. For exact-selector recall, BM25 obtains 0.455 {[}0.182, 0.727{]}, Rules Context 1.000 {[}1.000, 1.000{]}, Direct ECK 0.818 {[}0.545, 1.000{]}, and the no-evidence ablation 0.091 {[}0.000, 0.273{]}. For executable coverage, Direct ECK and Rules Context obtain 1.000 {[}1.000, 1.000{]}, while the no-evidence ablation obtains 0.636 {[}0.364, 0.909{]}.

We further compare against offline test-selection baselines that retrieve test files or test cases directly from the test suite without using ECK. Test-file BM25 reaches executable coverage 0.636 but exact selector recall 0.000; test-case BM25 obtains 0.091 on both metrics. This indicates that lexical test retrieval can identify a broad test file, but rarely recovers the precise executable selector. Explicit evidence provides this precision directly.

The 7B model shows the same qualitative pattern:

\begin{table*}[t]
\centering
\caption{Validation-planning results with the 7B model.}
\label{tab:eck-3}
\scriptsize
\setlength{\tabcolsep}{3pt}
\renewcommand{\arraystretch}{1.05}
\begin{tabularx}{\textwidth}{>{\raggedright\arraybackslash}X>{\raggedleft\arraybackslash}X>{\raggedleft\arraybackslash}X>{\raggedleft\arraybackslash}X}
\toprule
Method & File Recall & Exact Selector Recall* & Executable Coverage Recall* \\
\midrule
Plain & 0.904 & 0.000 & 0.273 \\
EKP Retrieval & 0.904 & 0.091 & 0.636 \\
Direct ECK & 0.962 & 0.727 & 0.727 \\
\bottomrule
\end{tabularx}
\end{table*}

\subsection{Critical Finding: ECK Complements Rather Than Replaces Retrieval}\label{critical-finding-eck-complements-rather-than-replaces-retrieval}

The data also show an important negative result. In the main 32B planning experiment, BM25 chunk retrieval and Direct ECK have identical file recall (0.981). We omit a symbol-level comparison because the pilot symbol gold mixed representation identifiers with code symbols. More broadly, lexical retrieval searches repository-wide implementation text, whereas direct ECK intentionally covers only a selected set of high-value units.

This finding changes the interpretation. ECK should not be evaluated as ``RAG but better.'' Its differentiating value is \textbf{validation-carrying context}: when a relevant ECKU exists, it can tell an agent what evidence to run, not just where code might be located.

This distinction appears at the task level. In six evidence-bearing tasks, BM25 locates relevant files but fails to produce a command covering the declared test, while Direct ECK produces covering commands. Direct ECK exactly recovers four of these six selector sets and emits broader covering class commands for \path{slugify_lowercase} and \path{slugify_separator}. This suggests that localization, executable coverage, and precise selector delivery are separable dimensions of agentic context quality.

\subsection{Critical Finding: Rules Can Carry Evidence, But Not Freshness or Impact}\label{critical-finding-rules-can-carry-evidence-but-not-freshness-or-impact}

The Rules Context baseline is intentionally strong: it renders the same ECK-derived hints as Markdown rules resembling project rules, memories, or an \path{AGENTS.md} file. Its perfect exact-selector recall shows that agents benefit from explicit evidence even without a structured ECK interface. This is important for industrial practice: rule files can be an effective delivery channel for validation hints.

However, the standard Rules Context does not provide the operational properties that ECK targets. Our Rules-with-IDs control shows that rules can carry identity strings, but they still have no executable source binding, patch-to-unit impact operation, validation state, or source/knowledge/contract/evidence freshness state. In our prototype, the same evidence can be rendered as rules for agent consumption, while the source-bound object remains the ECKU. This suggests a refined architecture: ECK should not replace rules or memories; it can generate them while preserving executable provenance.

We test this distinction with a staleness experiment. We generate a rules snapshot from ECKUs, then perturb executable source, agent-facing knowledge, or evidence while leaving the rules unchanged. Across 50 positive real-repository perturbations, ECK detects 50/50 stale cases while the rules snapshot detects 0/50 because it has no freshness mechanism; 17 unrelated same-file controls remain fresh. This result reframes the role of ECK: rules can deliver evidence to an agent, while ECK can determine whether the source-bound assertion has changed since that evidence last passed.

\subsection{RQ2: Deterministic Patch Evidence and Agent Projection Fidelity}\label{rq2-deterministic-patch-evidence-and-agent-projection-fidelity}

We first run deterministic patch impact analysis across 26 patches:

\begin{table*}[t]
\centering
\caption{Direct-ECK and extractive coverage across 26 patches.}
\label{tab:eck-4}
\scriptsize
\setlength{\tabcolsep}{3pt}
\renewcommand{\arraystretch}{1.05}
\begin{tabularx}{\textwidth}{>{\raggedright\arraybackslash}X>{\raggedleft\arraybackslash}X}
\toprule
Metric & Value \\
\midrule
Direct ECKUs & 17 \\
Direct ECKUs with Evidence & 16 \\
Extractive EKPs & 169 \\
Patches Touched by Direct ECK & 11 \\
Patches Touched by EKP-RAG & 26 \\
Direct Validation Plans & 11 \\
\bottomrule
\end{tabularx}
\end{table*}

Direct ECK has lower patch coverage than extractive EKP-RAG because it is selectively authored. This is expected. To evaluate the deterministic operation independently, we manually annotate direct ECKU impact from patch hunks and ECKU declarations without calling the impact engine. The gold set contains 11 positive patches and 12 patch-to-unit links across all 26 patches.

\begin{table*}[t]
\centering
\caption{Direct-impact accuracy against independently authored labels.}
\label{tab:eck-5}
\scriptsize
\setlength{\tabcolsep}{3pt}
\renewcommand{\arraystretch}{1.05}
\begin{tabularx}{\textwidth}{>{\raggedright\arraybackslash}X>{\raggedleft\arraybackslash}X}
\toprule
Direct-impact metric & Value \\
\midrule
True-positive unit links & 12 \\
False-positive unit links & 0 \\
False-negative unit links & 0 \\
Micro precision / recall / F1 & 1.000 / 1.000 / 1.000 \\
Exact-set accuracy, all patches & 1.000 \\
Exact-set accuracy, positive patches & 1.000 \\
\bottomrule
\end{tabularx}
\end{table*}

The final engine tracks exact added/deleted line positions inside each unified-diff hunk before intersecting them with AST-bounded ECKU spans. A whole-hunk prototype produced one false positive when a hunk touched the end of \path{slugify_params} and the start of \path{main}; exact changed-line tracking removes this boundary error. This evaluation supports direct source-span impact only, not relation-expanded consequences.

We then evaluate a model-backed patch-review workflow with Qwen2.5-Coder-32B-AWQ. Unlike the planning task, the model sees the actual patch diff and must identify affected ECK IDs, affected symbols, validation commands, and risk flags. Table 6 reports lenient parse results, with strict JSON compliance shown separately. As in RQ1, exact recall requires the declared selector, whereas coverage recall credits a broader executable command that contains it.

\begin{table*}[t]
\centering
\caption{Model consumption of patch evidence reports.}
\label{tab:eck-6}
\scriptsize
\setlength{\tabcolsep}{3pt}
\renewcommand{\arraystretch}{1.05}
\begin{tabularx}{\textwidth}{>{\raggedright\arraybackslash}X>{\raggedleft\arraybackslash}X>{\raggedleft\arraybackslash}X>{\raggedleft\arraybackslash}X>{\raggedleft\arraybackslash}X>{\raggedleft\arraybackslash}X>{\raggedleft\arraybackslash}X>{\raggedleft\arraybackslash}X}
\toprule
Method & Parsed & Strict JSON & Avg Prompt Chars & File Recall & ECK ID Recall & Exact Selector Recall & Executable Coverage Recall \\
\midrule
Plain Patch & 0.923 & 0.923 & 1344 & 0.923 & 0.000 & 0.000 & 0.273 \\
BM25 Patch & 0.962 & 0.962 & 7443 & 0.962 & 0.591 & 0.364 & 0.636 \\
Rules Patch & 0.923 & 0.923 & 4056 & 0.923 & 0.000 & 0.818 & 0.818 \\
Direct ECK Patch & 1.000 & 1.000 & 3875 & 1.000 & 1.000 & 1.000 & 1.000 \\
\bottomrule
\end{tabularx}
\end{table*}

ECK-ID and validation recall are computed over the 11 patches that touch at least one direct ECKU and have declared validation hooks. Bootstrap intervals over these 11 tasks are:

\begin{table*}[t]
\centering
\caption{Bootstrap intervals for ECK-ID and validation recall.}
\label{tab:eck-7}
\scriptsize
\setlength{\tabcolsep}{3pt}
\renewcommand{\arraystretch}{1.05}
\begin{tabularx}{\textwidth}{>{\raggedright\arraybackslash}X>{\raggedleft\arraybackslash}X>{\raggedleft\arraybackslash}X>{\raggedleft\arraybackslash}X}
\toprule
Method & ECK ID Recall 95\% CI & Exact Selector Recall 95\% CI & Executable Coverage Recall 95\% CI \\
\midrule
BM25 Patch & 0.591 {[}0.273, 0.864{]} & 0.364 {[}0.091, 0.636{]} & 0.636 {[}0.364, 0.909{]} \\
Direct ECK Patch & 1.000 {[}1.000, 1.000{]} & 1.000 {[}1.000, 1.000{]} & 1.000 {[}1.000, 1.000{]} \\
Plain Patch & 0.000 {[}0.000, 0.000{]} & 0.000 {[}0.000, 0.000{]} & 0.273 {[}0.000, 0.545{]} \\
Rules Patch & 0.000 {[}0.000, 0.000{]} & 0.818 {[}0.545, 1.000{]} & 0.818 {[}0.545, 1.000{]} \\
\bottomrule
\end{tabularx}
\end{table*}

The Direct ECK prompt contains the IDs and commands emitted by the same deterministic impact operation used to construct the corresponding targets. Its 1.000 values therefore measure \textbf{field-preserving consumption of a patch evidence report}, not independent impact discovery. Under that interpretation, the result sharpens a delivery distinction: rules transmit many validation hints, BM25 supplies broad source context at nearly twice the prompt length, and a typed ECK report gives the agent a compact named object whose declared evidence is preserved reliably. Independent impact accuracy is evaluated separately or left as a limitation; the model table must not be used as evidence for it.

\subsection{RQ3: Cross-Layer Projection Fidelity}\label{rq3-cross-layer-projection-fidelity}

The API-DB-business case study evaluates how faithfully agent-facing formats transport fields spanning endpoint payloads, service rules, database/entity fields, and tests. It contains 21 ECKUs across three business domains: orders/pricing, subscriptions/entitlements, and refunds/risk. We evaluate 18 patches, of which 17 directly touch an ECKU and one is a non-core model helper.

\begin{table*}[t]
\centering
\caption{Cross-layer projection fidelity.}
\label{tab:eck-8}
\scriptsize
\setlength{\tabcolsep}{3pt}
\renewcommand{\arraystretch}{1.05}
\begin{tabularx}{\textwidth}{>{\raggedright\arraybackslash}X>{\raggedleft\arraybackslash}X>{\raggedleft\arraybackslash}X>{\raggedleft\arraybackslash}X>{\raggedleft\arraybackslash}X>{\raggedleft\arraybackslash}X>{\raggedleft\arraybackslash}X>{\raggedleft\arraybackslash}X}
\toprule
Method & Avg Prompt Chars & ECK ID Recall & API Recall & DB Field Recall & Business Rule Recall & Exact Selector Recall & Executable Coverage Recall \\
\midrule
Plain & 1591 & 0.000 & 0.559 & 0.108 & 0.059 & 0.118 & 0.118 \\
BM25 & 9749 & 0.588 & 0.971 & 0.435 & 0.206 & 0.706 & 0.706 \\
Rules & 4947 & 0.000 & 0.971 & 0.617 & 0.471 & 1.000 & 1.000 \\
Rules with IDs & 5300 & 1.000 & 0.971 & 0.631 & 0.382 & 1.000 & 1.000 \\
Direct ECK & 6930 & 1.000 & 0.971 & 0.531 & 0.294 & 1.000 & 1.000 \\
ECK Projection & 4586 & 1.000 & 0.971 & 0.832 & 1.000 & 1.000 & 1.000 \\
\bottomrule
\end{tabularx}
\end{table*}

Bootstrap intervals over the 17 ECK-covered patches show the same pattern. ECK Projection obtains ECK ID recall 1.000 {[}1.000, 1.000{]}, business-rule recall 1.000 {[}1.000, 1.000{]}, exact-selector and executable-coverage recall 1.000 {[}1.000, 1.000{]}, and DB-field recall 0.832 {[}0.676, 0.971{]}. Rules match both validation metrics but have ECK ID recall 0.000 {[}0.000, 0.000{]} and business-rule recall 0.471 {[}0.353, 0.588{]}. BM25 has the longest prompts and lower ECK ID, DB-field, business-rule, and validation recall. Exact and coverage values happen to coincide in this case study because the predicted validation selectors are either exact or target unrelated scopes; they remain separately scored.

The Rules-with-IDs control is important: adding stable identifier strings to Markdown is sufficient for 1.000 identity, exact-selector, and executable-coverage recall. Identity visibility is therefore not unique to the ECK serialization. The remaining difference is operational: ECK is the source-bound object from which rules and projections are generated, validated, and invalidated.

As in RQ2, the cross-layer targets and ECK Projection fields are generated from the same ECKUs. The perfect business-rule result is therefore projection fidelity by construction, not evidence that ECK discovers all real cross-layer consequences. The useful design result is narrower: raw ECK is not necessarily the best delivery format; task-specific projection preserves schema-aligned fields more reliably and with fewer prompt characters. ECK remains the source-bound object, while rules or projections are delivery formats.

\subsection{RQ4: Freshness Perturbation}\label{rq4-freshness-perturbation}

We run two freshness studies. First, a controlled pricing demo tests four positive changes and three negative controls after validation.

\begin{table*}[t]
\centering
\caption{Controlled freshness perturbations.}
\label{tab:eck-9}
\scriptsize
\setlength{\tabcolsep}{3pt}
\renewcommand{\arraystretch}{1.05}
\begin{tabularx}{\textwidth}{>{\raggedright\arraybackslash}X>{\raggedleft\arraybackslash}X>{\raggedleft\arraybackslash}X}
\toprule
Perturbation & Expected & Observed \\
\midrule
source body & stale: source & stale: source \\
agent-facing knowledge & stale: knowledge & stale: knowledge \\
contract & stale: contract & stale: contract \\
evidence & stale: evidence & stale: evidence \\
unchanged & fresh & fresh \\
unrelated edit, same file & fresh & fresh \\
unrelated edit, other file & fresh & fresh \\
\bottomrule
\end{tabularx}
\end{table*}

All 7 cases are classified correctly. Source fingerprints use the AST-bounded function definition and body, excluding decorators; knowledge, contract, and evidence therefore produce their field-specific reasons without being conflated with source changes. A unit test additionally confirms that reordering otherwise identical relations does not change the normalized knowledge fingerprint.

Second, we run real-repository perturbations over the 17 direct ECKUs. For each unit we perturb the executable body and agent-facing intent, then add an unrelated edit in the same source file. For the 16 units with evidence, we also perturb the evidence command.

\begin{table*}[t]
\centering
\caption{Real-repository freshness perturbations.}
\label{tab:eck-10}
\scriptsize
\setlength{\tabcolsep}{3pt}
\renewcommand{\arraystretch}{1.05}
\begin{tabularx}{\textwidth}{>{\raggedright\arraybackslash}X>{\raggedleft\arraybackslash}X>{\raggedleft\arraybackslash}X>{\raggedleft\arraybackslash}X}
\toprule
Perturbation & Cases & Correct & Accuracy \\
\midrule
source body (positive) & 17 & 17 & 1.000 \\
agent-facing knowledge (positive) & 17 & 17 & 1.000 \\
evidence (positive) & 16 & 16 & 1.000 \\
unrelated same-file edit (negative) & 17 & 17 & 1.000 \\
total & 67 & 67 & 1.000 \\
\bottomrule
\end{tabularx}
\end{table*}

Across the 50 positive and 17 negative cases, sensitivity is 1.000, specificity is 1.000, and the false-positive rate is 0.000. This is a mechanism test over controlled perturbations, not a claim about naturally occurring developer edits or semantic equivalence. An earlier whole-file implementation failed the same-file controls; the AST-bounded span is necessary for the reported specificity, while \path{KnowledgeHash} closes the complementary blind spot created by excluding decorators from the source fingerprint.

\section{Discussion}\label{discussion}

\subsection{The Coverage-Authority Frontier}\label{the-coverage-authority-frontier}

The central empirical pattern is a frontier:

\begin{itemize}
\tightlist
\item
  Plain and extractive methods can provide broad context and sometimes strong localization.
\item
  Direct ECK has narrower coverage but stronger source binding and executable support when evidence matters.
\end{itemize}

This suggests a hybrid architecture. Use extractive methods for broad discovery and candidate generation. Use direct ECK for high-value units where agents need contracts, validation commands, and freshness checks. A mature system should allow extraction to propose candidate ECKUs, but require human review and executable evidence before those units become authoritative.

The patch-review experiment adds a second axis to this frontier: \textbf{auditability}. BM25 can identify many affected symbols, and rules can carry identities and validation hints, but ECK maintains the underlying object saying ``this patch directly overlaps unit \path{u}, whose declared evidence is \path{e}, whose freshness state is \path{v}.'' ECK makes patch review object-centric rather than text-centric. This is the main technical reason to represent selected knowledge with source binding rather than only externalizing it into rules, memories, summaries, or graphs.

\begin{table*}[t]
\centering
\caption{Capability comparison across context representations.}
\label{tab:eck-11}
\scriptsize
\setlength{\tabcolsep}{3pt}
\renewcommand{\arraystretch}{1.05}
\begin{tabularx}{\textwidth}{>{\raggedright\arraybackslash}X>{\raggedleft\arraybackslash}X>{\raggedleft\arraybackslash}X>{\raggedleft\arraybackslash}X>{\raggedleft\arraybackslash}X}
\toprule
Capability & Retrieval / Code Search & Rules / Memories & Direct ECK & ECK Projection \\
\midrule
Broad repository coverage & high & medium & selective & selective \\
Natural agent delivery & medium & high & medium & high \\
Stable code-unit identity & low & possible as generated text & high & high \\
Executable validation evidence & indirect & possible as text & first-class field & first-class field \\
Patch-to-knowledge impact & indirect & low & first-class operation & first-class operation \\
Source/knowledge/evidence freshness & low & low & first-class operation & first-class operation \\
Best role & discovery/localization & instruction delivery & source/evidence governance & schema-aligned delivery \\
\bottomrule
\end{tabularx}
\end{table*}

\subsection{Why Validation Planning Matters}\label{why-validation-planning-matters}

Most context work asks whether an agent can find relevant files or symbols. That is necessary but insufficient for safe software modification. An agent also needs to know what to validate. Our results show that even when plain or extractive context finds code, it often fails to recover validation commands. This matters for patch review, regression testing, and agent self-correction.

The planning and patch-review results also separate two mechanisms. Fresh rules can be an effective way to deliver validation commands to a model. Direct ECK adds a maintainability mechanism around those commands: evidence is source-bound, patch-impactable, and freshness-checkable. Thus ECK should be viewed as a source layer that can project into agent-facing rules, not as a replacement for every rule or memory system.

The Rules-with-IDs control confirms that generated ECK-like identifiers in Markdown recover identity as a text field. It does not by itself provide the underlying source span, evidence fingerprint, validation state, or stale-context detection. The core distinction is therefore not whether an identifier string can be shown to an agent, but whether that identifier is backed by a source-bound object with current executable support.

\subsection{Migration Path}\label{migration-path}

We envision five steps:

\begin{enumerate}
\def\labelenumi{\arabic{enumi}.}
\tightlist
\item
  Run extractive analysis to identify candidate knowledge-bearing functions.
\item
  Use an LLM to propose ECKU metadata.
\item
  Let developers accept or refine high-value ECKUs.
\item
  Run ECK validation in CI.
\item
  Feed ECK manifests and validation state to coding agents.
\end{enumerate}

The current prototype implements candidate suggestion and human-confirmed direct ECKUs for three repositories.

\subsection{Adoption and Authoring Cost}\label{adoption-and-authoring-cost}

ECK is intentionally selective. The expected adoption path is not to annotate every function, but to identify high-value units where agents repeatedly need business semantics, exact validation evidence, or patch-review provenance. In our prototype, extractive analysis and lexical search can propose candidate functions, an LLM can draft ECKU metadata, and a developer approves or edits the final unit before CI treats it as source-bound declared knowledge. The three evaluated repositories contain 17 human-confirmed ECKUs, 16 of which carry executable evidence. This scale is enough to cover parser invariants, CLI behavior, normalization rules, and public API contracts, while leaving ordinary helper code to retrieval and static analysis.

This suggests a practical maintenance model: extraction proposes coverage, direct ECK records a reviewed source-bound assertion, CI executes its declared evidence, and rules or memories are regenerated from the current ECK source. If an ECKU is not validated or becomes stale, agents should treat it as a warning-bearing object rather than trusted context.

\subsection{Artifact and Reproducibility}\label{artifact-and-reproducibility}

A focused reproducibility artifact packages the ECK CLI, tests, pricing example, patch evidence report, experiment scripts, gold labels, and saved raw model outputs separately from the paper workspace. The repository is pinned to the reviewed artifact snapshot and is available in the \href{https://anonymous.4open.science/r/eck-patch-evidence-0D5D/}{public artifact repository}.

The minimal CPU-only reproduction is:

\begin{lstlisting}
python -m pip install -e ".[dev]"
python -m eck report examples/eck_pricing \
  --patch examples/eck_pricing/patches/change_vip_discount.diff \
  --run-evidence --format markdown
python -m pytest
\end{lstlisting}

The expected report contains one directly impacted unit, \path{pricing.discount.vip_large_order}, one passing evidence item, and the outcome \path{verified}. The focused artifact currently has 11 passing tests and requires no GPU, network service, or proprietary dependency. The experiment bundle is organized as executable scripts and saved raw outputs:

\begin{table*}[t]
\centering
\caption{Artifact organization and reproduction entry points.}
\label{tab:eck-12}
\scriptsize
\setlength{\tabcolsep}{3pt}
\renewcommand{\arraystretch}{1.05}
\begin{tabularx}{\textwidth}{>{\raggedright\arraybackslash}X>{\raggedleft\arraybackslash}X}
\toprule
Purpose & Script or Artifact \\
\midrule
ECK unit discovery, query, validation, freshness, impact, report & \path{eck/*.py} \\
Planning experiment & \path{experiments/run_agent_context_experiment.py} \\
Patch-review experiment & \path{experiments/run_patch_review_experiment.py} \\
Patch-review reanalysis and bootstrap CI & \path{experiments/analyze_patch_review_results.py} \\
Rules staleness comparison & \path{experiments/compare_rules_staleness.py} \\
Positive/negative freshness controls & \path{experiments/run_eck_freshness_perturbation.py} and
\path{experiments/run_eck_real_repo_freshness.py} \\
Test-selection baselines & \path{experiments/compare_test_selection_baselines.py} \\
Result files & \path{experiments/results/*.json} and \path{experiments/results/*.md} \\
\bottomrule
\end{tabularx}
\end{table*}

All model-backed runs use temperature 0, top-k 6 context units or chunks, and an OpenAI-compatible endpoint. The reported 32B reruns use one NVIDIA H20 (97,871 MiB), NVIDIA driver 550.163.01, PyTorch 2.10.0+cu128, vLLM 0.19.1, and OpenAI Python client 2.44.0, with an 8,192-token model context limit and at most 800 generated tokens. The main model is \path{Qwen2.5-Coder-32B-Instruct-AWQ}; the secondary planning run uses \path{Qwen2.5-7B-Instruct}. Patch-review results are re-scored from saved raw model outputs, separating strict JSON compliance from lenient parseability.

\section{Threats to Validity}\label{threats-to-validity}

\textbf{Task construction.} Our initial pilot used patch-name-derived requests. The current experiment uses issue-style descriptions written independently from patch filenames, but they are still constructed from known patches. A stronger evaluation would use naturally occurring issue reports or human-written tasks created without inspecting the patch.

\textbf{Scale.} The real-repo evaluation covers three small Python libraries and 26 patches. The results show feasibility and a strong validation-planning signal, but not broad generality.

\textbf{Coverage.} Direct ECKUs cover selected high-value functions, not entire repositories. Lower patch coverage than extractive EKP-RAG is expected and should not be interpreted as a retrieval failure.

\textbf{Manual impact labels.} The direct-impact gold is independently authored from patch hunks and ECKU declarations rather than generated by the impact implementation, but it was produced by one annotator on controlled patches. A larger study should use multiple annotators, agreement reporting, naturally occurring commits, and separate labels for direct versus relation-expanded impact.

\textbf{Model dependence.} We evaluate two Qwen-family models. The qualitative validation-planning result is consistent across 7B and 32B, but broader model coverage would strengthen the claim.

\textbf{Freshness study.} Freshness perturbations are synthetic even when applied to real repositories. The 50 positive and 17 same-file negative cases test the sensitivity and specificity of AST-bounded source, normalized knowledge, and evidence fingerprints, not whether developers naturally maintain ECKUs correctly or whether all real edits are semantically relevant. A changed hash is a governance signal, not proof of a semantic change.

\textbf{Patch-review and projection construction.} The patch-review experiment gives the model the patch diff and evaluates report consumption, not whether the model generated the patch. More importantly, the Direct ECK targets and prompt fields share the deterministic impact generator, and the cross-layer targets and ECK Projection share ECKU fields. We therefore interpret their perfect values as projection fidelity, not independent impact discovery or end-to-end audit improvement.

\textbf{End-to-end repair.} We evaluate agent preparation and patch-review quality, not full patch generation success. The paper should not claim improved SWE-bench-style resolution without additional experiments.

\textbf{Responsible use.} Incorrectly authored or unvalidated ECKUs can mislead agents if treated as trusted context. ECK deployments should therefore surface freshness status, fail closed in CI for high-risk units, and distinguish validated evidence from unvalidated candidate metadata.

\subsection{Generative AI Use Disclosure}\label{generative-ai-use-disclosure}

Generative AI tools were used to assist with implementation, experiment scripting, literature discovery, language editing, and manuscript organization. The author defined the research questions and claims, selected and inspected the evaluated repositories and patches, executed and checked the experiments, independently authored the direct-impact gold labels, verified the cited sources, and takes responsibility for all content and results.

\section{Conclusion}\label{conclusion}

This paper argues that selected code units can directly express agent-usable knowledge rather than only being mined into external representations. We introduced ECKUs as source-bound, evidence-carrying, freshness-checkable code knowledge objects and implemented a Python prototype for authoring, querying, validating, checking direct patch impact, and projecting them into agent context. The paired evidence ablation shows that explicit declared commands primarily improve precise test-selector recovery, while broader executable coverage can sometimes be inferred from code or metadata. The projection studies show that typed reports and task-specific projections are reliable delivery formats, but do not independently validate impact discovery. The positive/negative freshness study shows that AST-bounded source fingerprints avoid same-file false positives and that a separate normalized knowledge fingerprint detects changes to agent-facing semantics and relations. The resulting design principle is practical and deliberately hybrid: use extraction and retrieval for coverage, rules or projections for delivery, and source-bound executable code knowledge for evidence and freshness governance.

\section{Data Availability Statement}\label{data-availability-statement}

The commit-pinned \href{https://anonymous.4open.science/r/eck-patch-evidence-0D5D/}{anonymous artifact repository} contains the ECK implementation, example, tests, patch-report reproduction, experiment scripts, gold labels, and saved raw model outputs described in Section 8. The minimal tool reproduction is CPU-only; rerunning the model-backed experiments requires the model environment reported in Section 8. No personal or human-subject data are used.

\balance
\bibliographystyle{ACM-Reference-Format}
\bibliography{references}

\end{document}